\documentclass[conference]{IEEEtran}
\IEEEoverridecommandlockouts
\usepackage{cite}
\usepackage{amsmath,amssymb,amsfonts}
\usepackage{algorithmic}
\usepackage{graphicx}
\usepackage{textcomp}
\usepackage{xcolor}
\usepackage{verbatim}
\def\BibTeX{{\rm B\kern-.05em{\sc i\kern-.025em b}\kern-.08em
    T\kern-.1667em\lower.7ex\hbox{E}\kern-.125emX}}

\usepackage{array, caption, floatrow, tabularx, makecell, booktabs}%
\usepackage{indentfirst}
\usepackage{graphicx}
\usepackage{graphics}
\usepackage{amsmath}
\usepackage{adjustbox}
\usepackage{footnote}   
\usepackage{array}
\usepackage{multirow}    
\usepackage[flushleft]{threeparttable}
\DeclareMathOperator{\E}{\mathbb{E}}
\usepackage{float}
\floatstyle{plaintop}
\restylefloat{table}

\begin{document}

\title{Uncertainty-based Visual Question Answering: Estimating Semantic Inconsistency \\between Image and Knowledge Base\\
\thanks{*This work was done when J. Chae was with Dongguk University.}
}

\author{
    \IEEEauthorblockN{Jinyeong Chae\IEEEauthorrefmark{1}}
    \IEEEauthorblockA{\textit{OKESTRO Co., Ltd} \\
    Seoul, Republic of Korea \\
    jiny491@gmail.com}
    \and 
    \IEEEauthorblockN{Jihie Kim}
    \IEEEauthorblockA{\textit{Dept. of Artificial Intelligence} \\
    \textit{Dongguk University} \\
    Seoul, Republic of Korea \\
    jihie.kim@dgu.edu}
}

\maketitle

\begin{abstract}


Knowledge-based visual question answering (KVQA) task aims to answer questions that require additional external knowledge as well as an understanding of images and questions. Recent studies on KVQA inject an external knowledge in a multi-modal form, and as more knowledge is used, irrelevant information may be added and can confuse the question answering. In order to properly use the knowledge, this study proposes the following: 1) we introduce a novel semantic inconsistency measure computed from caption uncertainty and semantic similarity; 2) we suggest a new external knowledge assimilation method based on the semantic inconsistency measure and apply it to integrate explicit knowledge and implicit knowledge for KVQA; 3) the proposed method is evaluated with the OK-VQA dataset and achieves the state-of-the-art performance.

\end{abstract}


\section{Introduction}
Knowledge-based visual question answering (KVQA) task is to answer questions that require an understanding of images, questions, and additional external knowledge. The KVQA task is proposed with the aim of reaching human-level reasoning. Injecting huge knowledge related to the entities identified from images and questions in a multi-modal form is among the tasks being researched. However, as the knowledge base (KB) is often incomplete, when the context of the entities is not fully consistent with the KB, irrelevant information can be retrieved and confuse the question answering. 
\begin{figure}[ht]
    \centering
    \includegraphics[width=9cm]{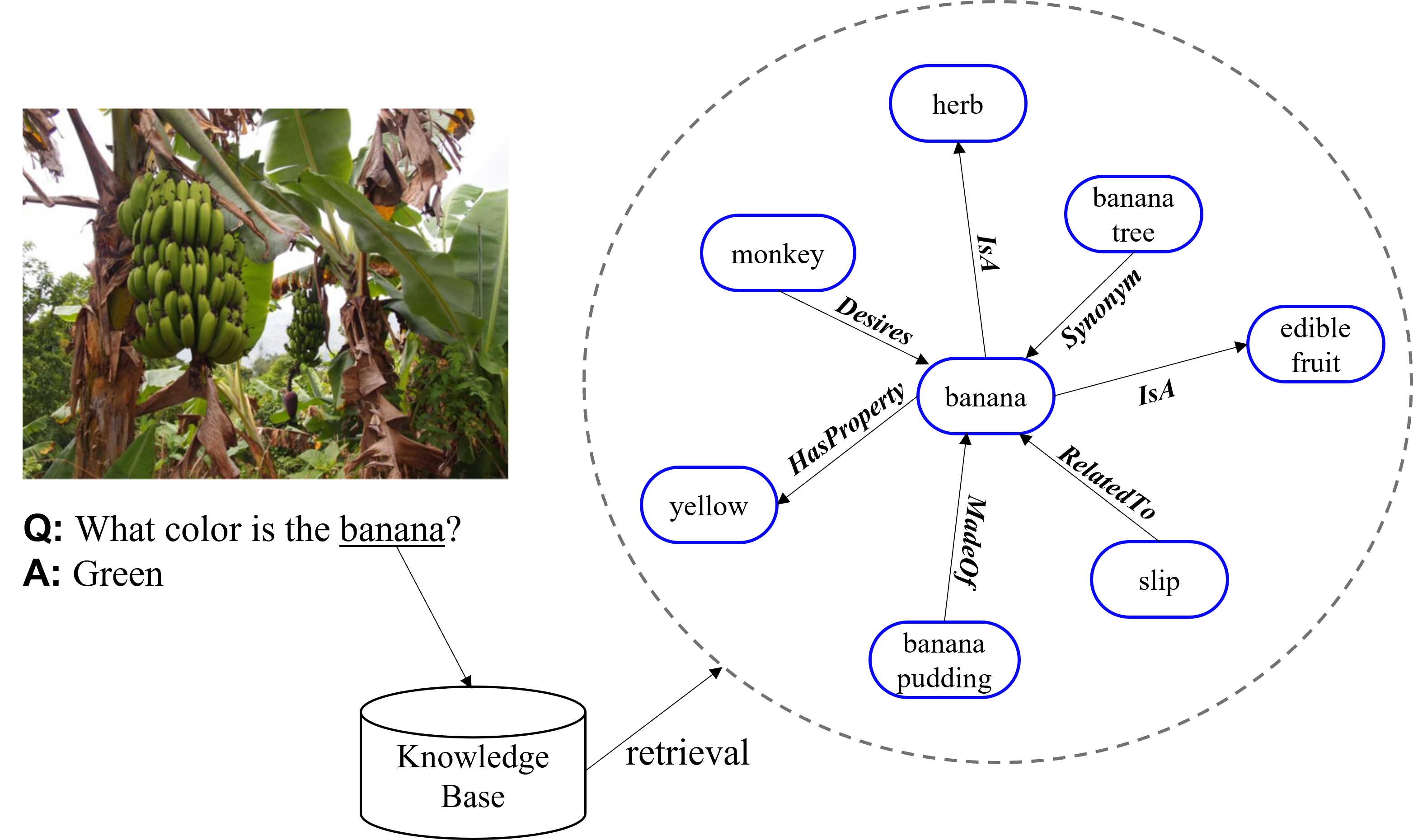}
    \caption{An example of visual question answering, occurring a semantic inconsistency between the image and external knowledge. The knowledge graph is the external knowledge extracted according to the word of the question.}
    \label{fig:dataset sample}
\end{figure}

For the example in Fig. \ref{fig:dataset sample}, the question can be answered with a full understanding of the image and question. However, the predicted answer can become yellow when we use related general knowledge, i.e., \textit{(banana, HasProperty, yellow)}. In this case, there is a conflict between the image and the knowledge base. We define semantic inconsistency as such conflicts between the image context and the knowledge extracted based on the object in the image or question keywords.
For KVQA, there have been a lot of approaches introduced to make use of external KB with the given image and the question. Recent studies \cite{RN9} and \cite{RN29} suggested a method of extracting external knowledge by using the object keywords of the image and the words in the question. However, as shown above, these approaches can suffer from semantic inconsistency when the context information of the VQA is not well used given the new information from the KB. \cite{RN14}, \cite{RN10}, and \cite{RN2} introduced graph-based approaches for the KVQA where the work focuses on how to extract the needed knowledge using graph algorithms. However, they lack considering how well the extracted knowledge match with the given image and the question, and the extracted knowledge can rather confuse the answer prediction. In such cases, we believe it is necessary to evaluate and adjust the amount of the external knowledge injected based on semantic inconsistency between the context of the image and the knowledge. In estimating semantic inconsistency, first we propose to make use of caption generation results which can indicate unusuality of the image. In addition, we develop a new uncertainty-based measure that uses knowledge context model pre-trained with commonsense knowledge. When generating the knowledge context for an image, the context that is inconsistent with the actual image context may incur uncertainty, and the semantic inconsistency can be estimated through an  uncertainty measure. Furthermore, the inconsistency can be also estimated by the similarity between the generated knowledge context and the image context. This study combines these into a new approach for measuring such inconsistencies and introduces a new way of assimilating external knowledge. 
This study is summarized as follows:  
\begin{itemize}
\item We introduce a new semantic inconsistency measure based on caption generation, which is an ensemble of a) uncertainty of the caption and b) similarity between the caption generated with the KB and the ground-truth caption
\item We propose an external knowledge assimilation method based on the proposed semantic inconsistency measure to control the use of external knowledge in KVQA.
\item We apply the proposed method for combining explicit and implicit knowledge passed through Relational Graph Convolution Networks (RGCN) and VisualBERT, respectively in KVQA and achieve the state-of-the-art result when evaluated with the OK-VQA dataset.
\end{itemize}

\section{Related work} \label{rw}
\subsection{KVQA approaches using pre-trained model}
A lot of researches have studied image and text as a multi-modal form. By tokenizing the object in an image, an alignment between an object and text has been proposed to apply a self-attention model \cite{RN15}\cite{RN24}. In addition, \cite{RN24} showed such models achieve better performance in various downstream tasks compared with other vision-language approaches \cite{RN23}. Therefore, this study experiments with the approach suggested by \cite{RN24} for extracting the implicit knowledge. Multi-modal approaches using image features from Faster R-CNN or ResNet and question embedding of pre-trained models are also proposed \cite{RN8}\cite{RN12}. \cite{RN8} generated joint representation through Bilinear Attention Map. \cite{RN12} extracted image-text joint representation by using image features and question embedding, and proposed a 3-way Tucker fusion method. In addition to using pre-trained models, there have also been studies trying to solve VQA tasks using additional external knowledge. \cite{RN9} proposed ArticleNet using Wikipedia search API related to keywords of an image and words of the question. A method for extracting an external knowledge related to the objects in an image was also introduced \cite{RN29}. \cite{RN29} extracted the knowledge by using the object label output from the Faster R-CNN model. This study extracts more relevant knowledge by using not only image object keywords, but also words in the question.

\subsection{Graph-based KVQA approaches}
Besides using pre-trained models, studies using graph-based models were proposed \cite{Hudson}\cite{heterograph}\cite{Noa}\cite{RN40}\cite{RN14}. \cite{Hudson} suggested the Neural State Machine based on a probabilistic graph for reasoning on VQA. \cite{Noa} introduced a video scene graph and caption generation method, and applied them for reasoning on video-QA task. \cite{heterograph} studied a heterogeneous graph alignment network considering inter-and intra-modality for video-QA. \cite{RN40} proposed a method to create graphs from visual, linguistic, and numeric features and suggested an aggregator that combines the features. However, because the study focuses on the contents of the image, the method has a limitation in answering a question that requires additional knowledge. \cite{RN14} suggested graph-based VQA for capturing the interrelationship between objects and entities of external knowledge by combining concept graph and scene graph. However, the scene graph relation is limited because only locational information is considered, and in the OK-VQA dataset, the extraction method for location-based scene graphs does not show significant performance improvements. 

Moreover, studies using a pre-trained model and graph-based model have been suggested \cite{saqur}\cite{RN10}\cite{RN2}. \cite{saqur} introduced multi-modal graph networks for compositional generalization in VQA, but the method is evaluated with the VQA task that only requires object detection or recognition in answering questions for the object shape and the number of objects. \cite{RN10} proposed a Knowledge Graph Augmented model using a pre-trained object detection model and graph-based method. However, the knowledge subgraph is generated by using the image object labels and the words of the question without considering the image-question context. \cite{RN2} proposed to integrate image-text representation from the BERT-based model and graph information based on the concept of image objects and questions. However, when there are conflicts between the graph and the pre-trained model representation, use of knowledge can hinder the question answering, as described above. This study proposes a new approach that measures semantic inconsistencies between KB and the given problem, and moderates the use of knowledge based on the measurement. 

\section{Approach}
This section introduces a semantic inconsistency measure that makes use of uncertainty and semantic similarity modeling. 
\subsection{Semantic inconsistency between an image and an external KB} \label{uem}
In this study, we utilize caption generation to measure semantic inconsistency between an image and external KB. Inspired by \cite{RN1}, we adopt uncertainty model of caption generation and introduce a novel measure for estimating semantic inconsistency between the KB and the VQA context. 
\subsubsection{Ensemble-based uncertainty estimation for KVQA}
In the existing image captioning, to generate a sentence $y$ when an input $x$ is given, the conditional distribution $p(y|x)$ is learned and tokens are continuously predicted from an autoregressive distribution. 

\begin{equation} \label{eqn:1}
p(y|x)= p(y_1|x)\prod_{i=2}^{k}p(y_i|x, y_1, \cdots , y_{i-1})
\end{equation}

In Eq. (\ref{eqn:1}), $y_i$ denotes the token corresponding to the index $i$ in sentence $y$, and the given set $\{x, y_1, \cdots, y_{i-1}\}$ denotes context $c_i$ for predicting the token corresponding to $i$. The number of tokens that can be predicted is limited based on the given context. For example, the word ``beach" cannot be generated when an image of a cat on a desk is given. When a set of words irrelevant to the context is denoted hallucinated word $V_h^{(c_i)}$, the following equation can be written

\begin{equation}\label{eqn:2}
p(y_i\in V_{h}^{(c_i)})=\sum_{v\in V_{h}^{(c_i)}}^{}p(y_i=v|c_i)
\end{equation}

In image captioning, token prediction in a given context is calculated with the following cross-entropy equation. The equation can be divided into two based on an entropy of the set of words relevant to the context and that of the set of words irrelevant to the context as

\begin{equation}\label{eqn:3}
\begin{aligned}
H(y_i|c_i) = -\sum_{v\in V}p(y_i=v|c_i)logp(y_i=v|c_i) \\
=-\sum_{v\in V \setminus V_{h}^{(c_i)}}^{}p(y_i=v|c_i)logp(y_i=v|c_i) \\
-\sum_{v\in V_{h}^{(c_i)}}^{}p(y_i=v|c_i)logp(y_i=v|c_i)
\end{aligned}
\end{equation}

The uncertainty that can be predicted by the Eq. (\ref{eqn:3}) can be divided into two: 1) uncertainty that appears in selection of a token that describes the context; 2) uncertainty that appears due to the interference of words irrelevant to the context or an insufficient training system. The latter is directly related to calculating hallucinated words that are irrelevant to the given context. We make use of the latter in measuring uncertainty in KVQA, as described below. The latter can be decomposed into two: aleatoric uncertainty and epistemic uncertainty \cite{Armen}\cite{Depeweg}\cite{Gal}. The uncertainties can be measured by an ensemble-based model \cite{Balaji} and calculated as follows:
\begin{equation}\label{eqn:4}
\begin{aligned}
u_{al}(y_i|c_i) = \E_{q(w)}[H(y_i|c_i, w)] \\
= \frac{1}{M}\sum_{m=1}^{M}H_m(y_i|c_i)
\end{aligned}
\end{equation}

\begin{equation}\label{eqn:5}
\begin{aligned}
u_{ep}(y_i|c_i) = H(y_i|c_i) - \E_{q(w)}[H(y_i|c_i, w)] \\
= H(y_i|c_i) - u_{al}(y_i|c_i)
\end{aligned}
\end{equation}

In Eq. (\ref{eqn:4}), $w$ denotes the model weights and $q(w)$ denotes the posterior distribution of weights in the training data. If the weights are fixed, $H(y_i|c_i, w)$ represents the uncertainty related to the data. Aleatoric uncertainty can be written as $\E_{q(w)}[H(y_i|c_i, w)]$ and calculated by the mean of $H_m(y_i|c_i)$. Epistemic uncertainty can also be written by the difference between the entropy $H(y_i|c_i)$ of $p(y_i|c_i)$ and aleatoric uncertainty in Eq. (\ref{eqn:5}). 

\begin{figure*}
    \centering
    \includegraphics[width=17cm]{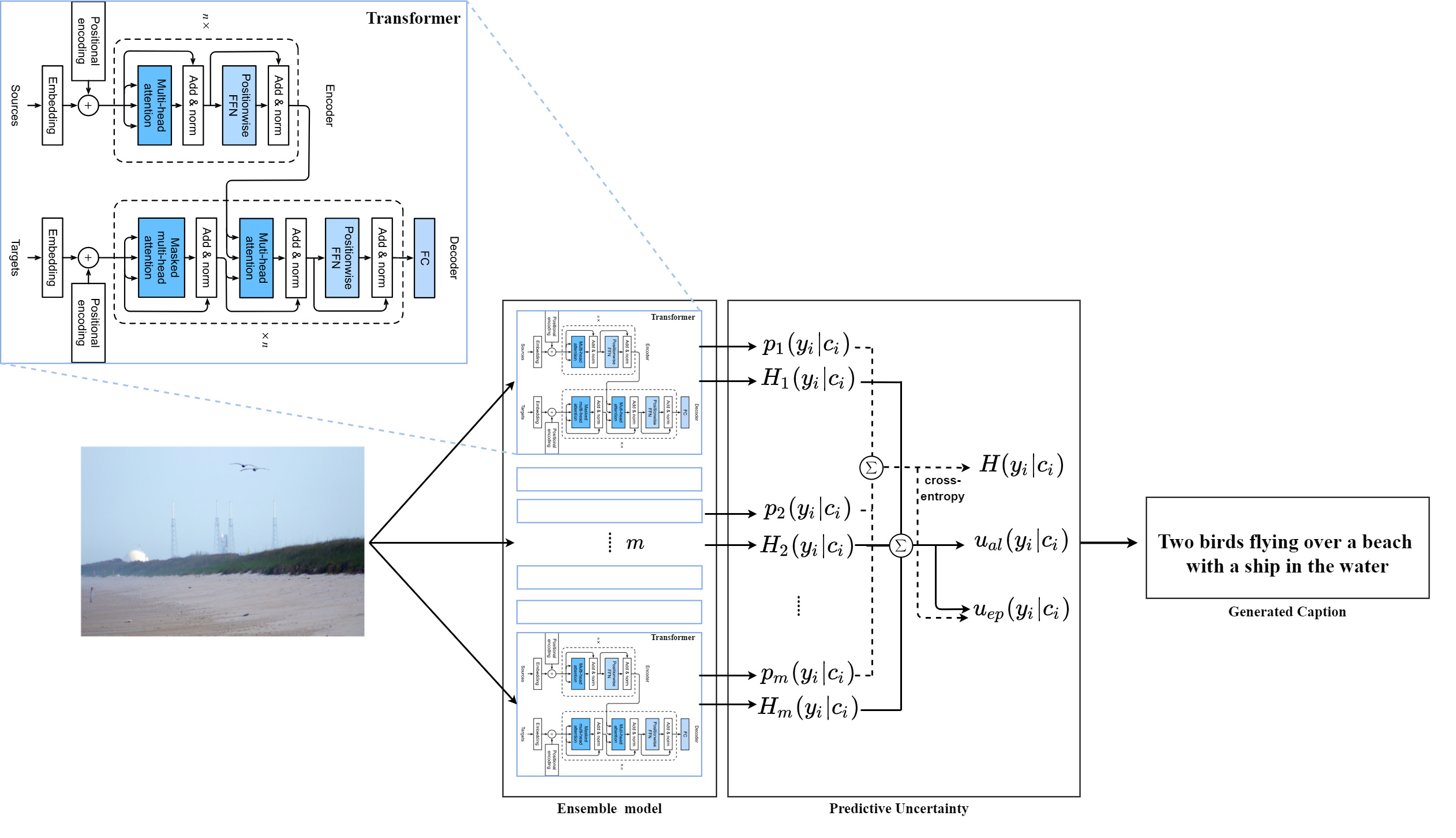}
    \caption{Ensemble-based uncertainty estimation based on caption generation. The given image shows birds flying over a beach.}
    \label{fig:gen}
\end{figure*}
A recent study shows that the model pre-trained with a large amount of image captioning data incorporates commonsense knowledge that is implicit in the data \cite{vlbert}. We use such a pre-trained model (with commonsense knowledge) to generate captions including knowledge context from the KVQA image data, and predict the uncertainty of the knowledge for the given VQA using the above ensemble model. The proposed method is illustrated in Fig. \ref{fig:gen}. As shown in Fig \ref{fig:gen}, when the image that birds are flying over the sand beach is given, the generated caption with commonsense knowledge is that two birds flying over a beach with a ship in the water. The caption reflects a general knowledge that ships are on a beach.
\subsubsection{Measuring similarity between caption sentences}
In addition to the above uncertainty model, this study proposes a novel measure that predicts the uncertainty of the knowledge based on the similarity between the generated and the ground-truth caption. That is, if the generated caption with the pre-trained model is much different from the ground-truth caption, the generated commonsense knowledge may be not much of use for the given problem. The S-BERT sentence embedding method \cite{RN22} is used to calculate the caption similarity. The similarity between the caption embeddings is calculated as follows:

\begin{equation}\label{eqn:6}
\begin{aligned}
sim^{cap}(S_g, S_t) = \frac{f(S_g)\cdot f(S_t)}{\left \| f(S_g) \right \| \cdot \left \| f(S_t) \right \| }, f:encoder
\end{aligned}
\end{equation}

In Eq. (\ref{eqn:6}), $S_g$ and $S_t$ denote the generated caption and the ground-truth caption, respectively. $f$ is an encoder for extracting a representation. The similarity is calculated from dot product between the representations of the generated caption and the ground-truth.
\subsection{Knowledge-based visual question answering} \label{kvqa}
Based on the above uncertainty measures, we present a new approach that integrates implicit knowledge and explicit knowledge external KB into KVQA.  
\subsubsection{Use of knowledge based on semantic consistency}

\begin{figure}
    \centering
    \includegraphics[width=9cm]{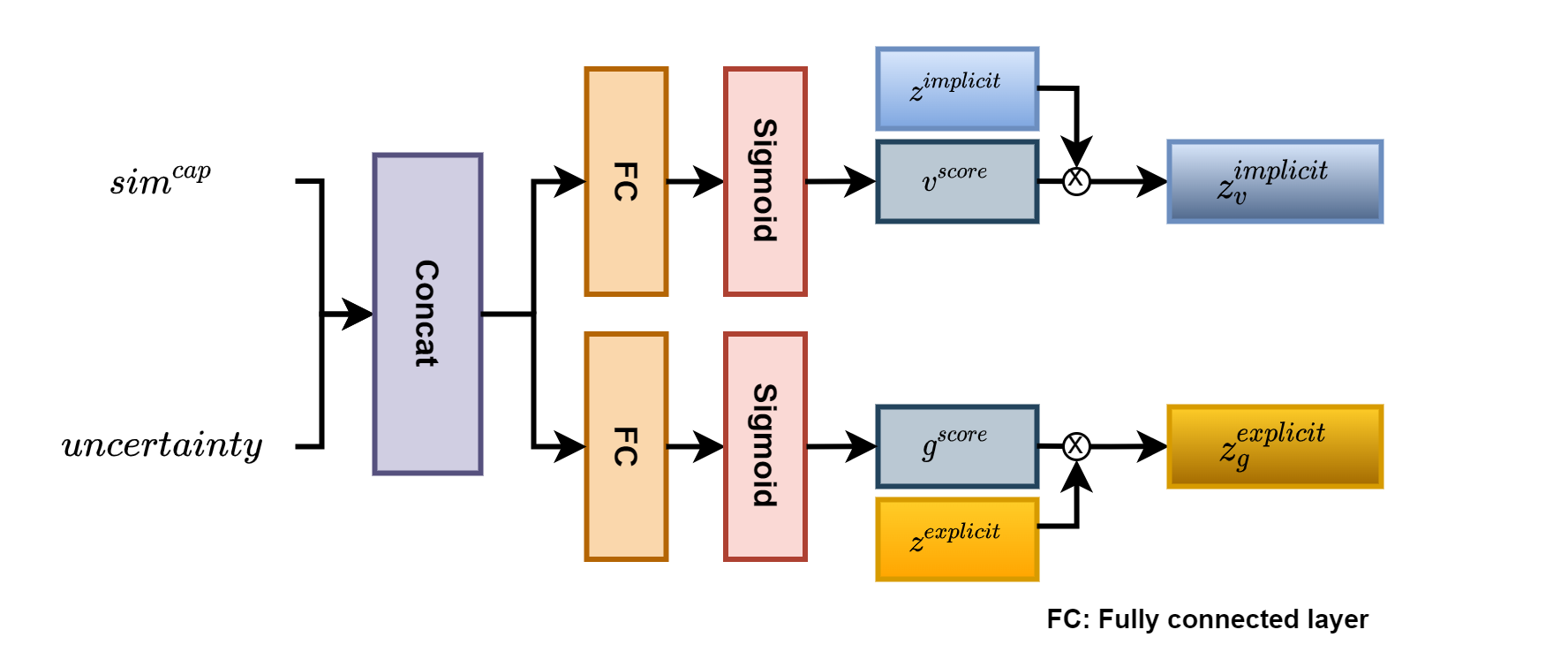}
    \caption{Use of knowledge adjusted based on uncertainty measures.}
    \label{fig:cm}
\end{figure}

As shown in Fig. \ref{fig:cm}, we adjust the use of the given KB based on the above mentioned uncertainty measures. In KVQA, when the uncertainty is high and the similarity is low, the scores of the metrics mean that the meaning of generated caption with commonsense knowledge is far from that of the image. Therefore, the system tends to attend to the content of image-question information. Otherwise, the external knowledge is more attended. 

\begin{equation}\label{eqn:7}
\begin{aligned}
v^{score} = \sigma(W_v * [sim^{cap}, u]) \\ 
g^{score} = \sigma(W_g * [sim^{cap}, u]) \\ 
\mathbf{z}_v^{implicit} = v^{score} * \mathbf{z}^{implicit} \\ 
\mathbf{z}_g^{explicit} = g^{score} * \mathbf{z}^{explicit}
\end{aligned}
\end{equation}

As shown in Fig. \ref{fig:cm} and Eq. (\ref{eqn:7}), $v^{score}\in\mathbb{R}$ and $g^{score}\in\mathbb{R}$ are calculated through a fully connected layer and a sigmoid function $\sigma$ after concatenating the similarity $sim^{cap}\in\mathbb{R}$ and the uncertainty $u\in\mathbb{R}$. Each of the score values is finally represented by $\mathbf{z}_v^{implicit}$ and $\mathbf{z}_g^{explicit}$ through a dot product between the pre-calculated representation $\mathbf{z}^{implicit}\in\mathbb{R}^{d_{zi}}$ extracted from a vision-language model and the knowledge representation $\mathbf{z}^{explicit}\in\mathbb{R}^{d_{ze}}$, where $\mathbf{z}^{explicit}$ is extracted from a knowledge graph. The use of image-question information and KB are adjusted based on the score. 

\subsubsection{KVQA with semantic consistency model}
For KVQA, this study proposes a semantic consistency model that relies on the uncertainty measures described above. The model relies on two types of knowledge sources inspired by \cite{RN2}: 1) explicit knowledge and 2) implicit knowledge. The former is the knowledge extracted from RGCN that has the external KB as input \cite{rgcn}. The latter is a vision-language embedding extracted from VisualBERT trained with a large-scale data. Furthermore, the use of the explicit and implicit knowledge is adjusted based on the uncertainty estimation, as described in section \ref{uem}.

\begin{figure*}
    \centering
    \includegraphics[width=13cm]{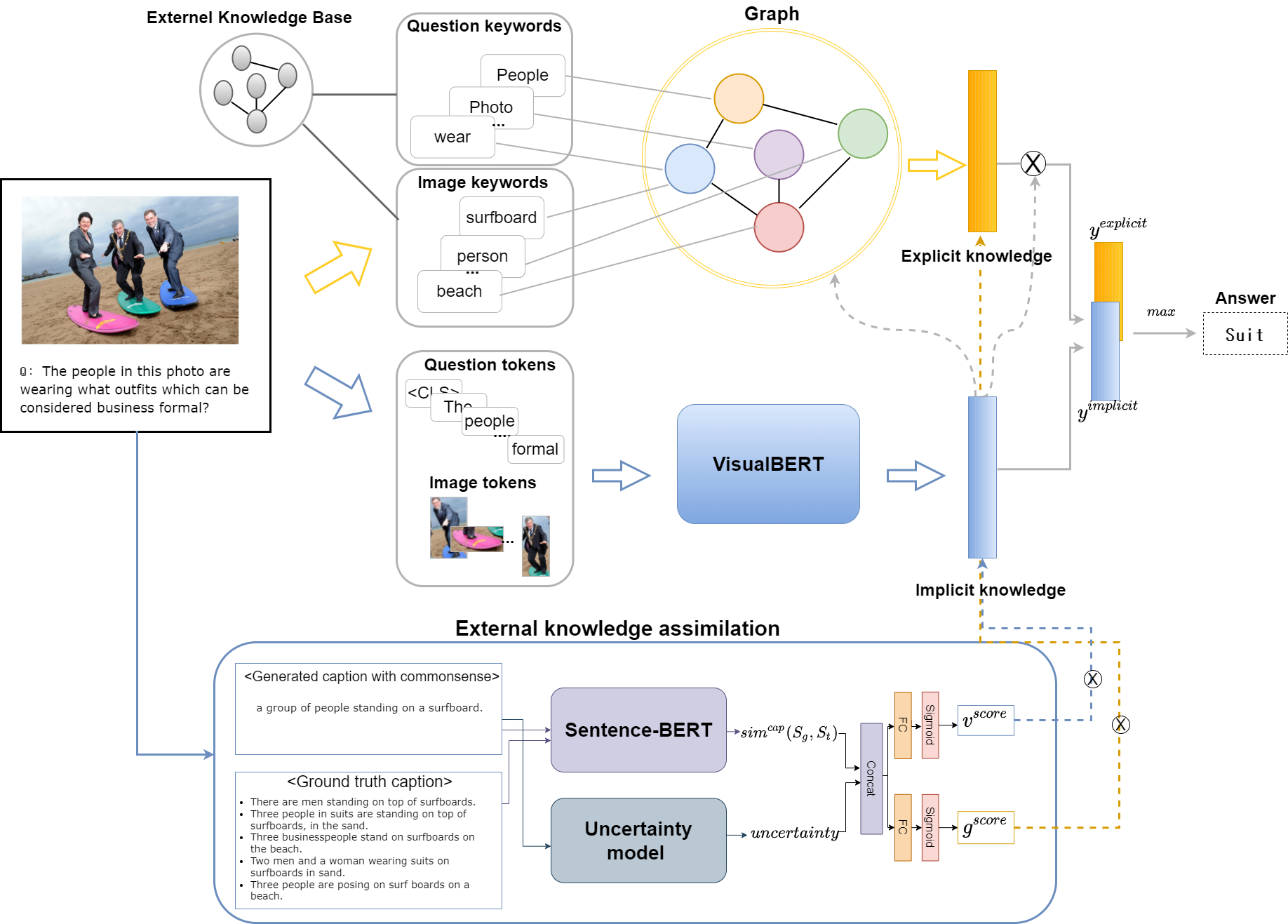}
    \caption{Overall model architecture. In the external knowledge assimilation, when the image is given, the generated caption with commonsense is a group of people standing on a surfboard. The five ground-truth captions of the image are as follows: 1) there are men standing on top of surfboards; 2) three people in suits are standing on top of surfboards, in the sand; 3) three business people stand on surfboards on the beach; 4) two men and a woman wearing suits on surfboards in sand; 5) three people are posing on surfboards on a beach. A final answer is predicted with explicit knowledge and implicit knowledge combining the external knowledge assimilation.}
    \label{fig:model}
\end{figure*}

\textbf{Explicit knowledge extraction}: Explicit knowledge is created by extracting relevant knowledge from the external KB, using the objects recognized in the image. In this study, about 4000 image keywords including objects, places, and attributes of objects are extracted with the following models: 1) ResNet-152 (ImageNet  \cite{RN55}); 2) ResNet-18 (Place365 \cite{RN56}); 3) Faster R-CNN (VisualGenome \cite{RN58}); 4) Mask-RCNN (LVIS \cite{RN57}). External KB used are as follows: 1) DBPedia (categorical information) \cite{dbpedia}; 2) ConceptNet (commonsense knowledge) \cite{conceptnet}; 3) VisualGenome (spatial relationship) \cite{RN58}; 4) hasPartKB (part relationship) \cite{haspartkb}. The relevant knowledge is retrieved with image keywords and question words. As a result, a total of 36,000 edges and 8,000 nodes are extracted. For integrating knowledge graphs, we use RGCN that distinguishes types and directions of edges in this study. The followings are used as RGCN inputs: 1) keyword presence that indicates words in the question with filtered words with one-hot matrix; 2) an image keyword probability extracted from a pre-trained model; 3) Word2vec representation of each keyword or average Word2vec representation of multiple words \cite{word2vec}; 4) implicit knowledge representation $\mathbf{z}^{implicit}$ extracted from VisualBERT. The extracted explicit and implicit knowledge are integrated into KVQA as described above.  

\textbf{Implicit knowledge extraction}: Transformer-based language models trained with a large-scale corpus are known to learn commonsense. Therefore, we use the VisualBERT model to make use of the implicit knowledge generated from the image and the question \cite{RN24}, as shown in Fig. \ref{fig:model}. Although there are various studies that align images and sentences together, we apply the appropriate model to our task using experiments in \cite{RN23}. The question representations are extracted by the pre-trained BERT model with BookCorpus dataset and English Wikipedia, and we use the representations as the input to the VisualBERT model. Furthermore, the visual representations are extracted from the Faster R-CNN model pre-trained with VisualGenome/COCO dataset and the result becomes the input of VisualBERT. To produce $\mathbf{z}^{implicit}$ representation, we use mean-pooling with outputs extracted from the VisualBERT model.

To get a final answer, we predict the answer within a set of vocabulary of answers $V \in \mathbb{R}^{v}$ where $v$ is the size of vocabulary. The final implicit score $y^{implicit}$ and explicit score $y_i^{explicit}$ are calculated to predict the answer from the set $V \in \mathbb{R}^{v}$ as follows. 

\begin{equation}\label{eqn:11}
\begin{aligned}
y^{implicit} = \sigma(W * \mathbf{z}_v^{implicit} + b)
\end{aligned}
\end{equation}
\begin{equation}\label{eqn:12}
\begin{aligned}
y_i^{explicit} = \sigma((W_{ge} * \mathbf{z}_{i, g}^{explicit} + b_{ge})^T\\(W_{vi}*\mathbf{z}_v^{implicit} + b_{vi}))
\end{aligned}
\end{equation}

In Eq. (\ref{eqn:11})-(\ref{eqn:12}), $y^{implicit}$ is calculated with a fully connected layer with weight $W$ and bias $b$, and a sigmoid function. In addition, $y_i^{explicit}$ is a score of word \(i\) corresponding to \(V \in \mathbb{R}^{v}\), computed with linear transformations of $\mathbf{z}_{i, g}^{explicit}$ and $\mathbf{z}_v^{implicit}$. The final answer is selected by choosing the highest value from both \(y^{implicit}\) and \(y^{explicit}\) . The model is trained with binary cross-entropy. 

\section{Experiments and results} \label{exp} 
\subsection{Dataset and baseline}
We use the OK-VQA dataset \cite{RN9} which is a popular KVQA benchmark dataset. The dataset consists of a total of 14,031 images and 14,055 questions. The detailed dataset sizes for training and testing are shown in Table \ref{tab:6}. For the validation dataset, we also use 1/3 of the training dataset based on the number of questions.

\begin{table}
\caption{Table of OK-VQA dataset.} \label{tab:6}
\begin{center}
\begin{adjustbox}{width=5cm,center}
\begin{tabular}{ccc}
\toprule
\multicolumn{1}{c}{\bf Dataset}  &\multicolumn{1}{c}{\bf \# of images}  &\multicolumn{1}{c}{\bf \# of questions} \\ \toprule
Train     &  8,998   & 9,009  \\ \midrule
Test & 5,033    & 5,046 \\ \midrule
Total & 14,031 & 14,055 \\ \bottomrule
\end{tabular}
\end{adjustbox}
\end{center}
\end{table}

\begin{table}
\caption{Table of MSCOCO dataset.} \label{tab:1}
\begin{center}
\begin{adjustbox}{width=5cm,center}
\begin{tabular}{ccc}
\toprule
\multicolumn{1}{c}{\bf Dataset}  &\multicolumn{1}{c}{\bf \# of images}  &\multicolumn{1}{c}{\bf \# of captions} \\ \toprule
Train   &   82,783    &   413,915  \\ \midrule
Validation  &   40,504  &   202,520 \\ \midrule
Test    &   40,775  &   379,249  \\ \midrule
Total   &   164,062 &   995,684 \\ \bottomrule

\end{tabular}
\end{adjustbox}
\end{center}
\end{table}

MSCOCO dataset \cite{RN16} is used to pre-train baseline models that generate captions. The dataset size is shown in Table \ref{tab:1}. In addition, Att2in \cite{RN18}, BuDn \cite{RN19}, and Transformer \cite{RN20} are selected as the baseline models for caption generation, which are the representative image captioning models, and are used to generate captions of the OK-VQA dataset. 

\subsection{Metrics}
In this study, a standard evaluation metric used in VQA challenge \cite{RN3} is employed to evaluate the performance with the OK-VQA dataset. Furthermore, we evaluate the generated caption with BLEU \cite{RN4}, CIDER \cite{RN6}, METEOR \cite{RN7}, and ROUGE-L \cite{RN5} metrics.  

\subsection{Uncertainty-based caption generation}

\begin{table}
\caption{Table of pearson correlation with the uncertainty and the similarity. $sim^{cap}$ represents a caption similarity. $un^{al}$ and $un^{ep}$ represent aleatoric uncertainty and epistemic uncertainty, respectively.}\label{tab:3}
\begin{center}
\begin{adjustbox}{width=3.5cm,center}
\begin{tabular}{cc}
\toprule
\multicolumn{1}{c}{\bf }  &\multicolumn{1}{c}{\bf Corr}  \\ \toprule
$sim^{cap}$ \& $un^{al}$    &  -0.1907   \\ \midrule
$sim^{cap}$ \& $un^{ep}$       &  -0.1653  \\ \midrule
$un^{al}$ \& $un^{ep}$    &  0.4518        \\\bottomrule
\end{tabular}
\end{adjustbox}
\end{center}
\end{table}

\begin{table*}
\caption{Performances of image captioning with commonsense knowledge on the OK-VQA dataset.}
\label{tab:2}
\begin{center}
\begin{adjustbox}{width=16cm, height=1cm, center}
\begin{tabular}{llllllll}
\toprule
\multicolumn{1}{c}{\bf Model}  &\multicolumn{1}{c}{\bf BLEU-1}  &\multicolumn{1}{c}{\bf BLEU-2}  &\multicolumn{1}{c}{\bf BLEU-3}  &\multicolumn{1}{c}{\bf BLEU-4}  &\multicolumn{1}{c}{\bf CIDER} &\multicolumn{1}{c}{\bf METEOR} &\multicolumn{1}{c}{\bf ROUGE-L} \\ \toprule
Att2in \cite{RN18}    & 0.7843±0.00005    & 0.6077±0.0002     & 0.4508±0.00032   & 0.3302±0.00038    & 1.0833+0.0016    & 0.2604±0.00018    & 0.5561±0.00019 \\ \midrule
BuDn \cite{RN19}        &  0.8123±0.00015    & 0.6516±0.00009      & 0.5017±0.00003  & 0.3786±0.00004    & 1.2527±0.00039    & 0.2858±0.00002 & 0.5859±0.00005 \\ \midrule
Transformer \cite{RN20} & {\bf 0.8290±0.00028}    & {\bf 0.6828±0.00036}      & {\bf0.5410±0.0004}   & {\bf0.4216±0.0004}    & {\bf1.3864±0.0012}     & {\bf 0.2997±0.00013}     & {\bf 0.6043±0.00023}  \\ \bottomrule
\end{tabular}
\end{adjustbox}
\end{center}
\end{table*}

\begin{figure*}[ht]
    \centering
    \includegraphics[width=14cm]{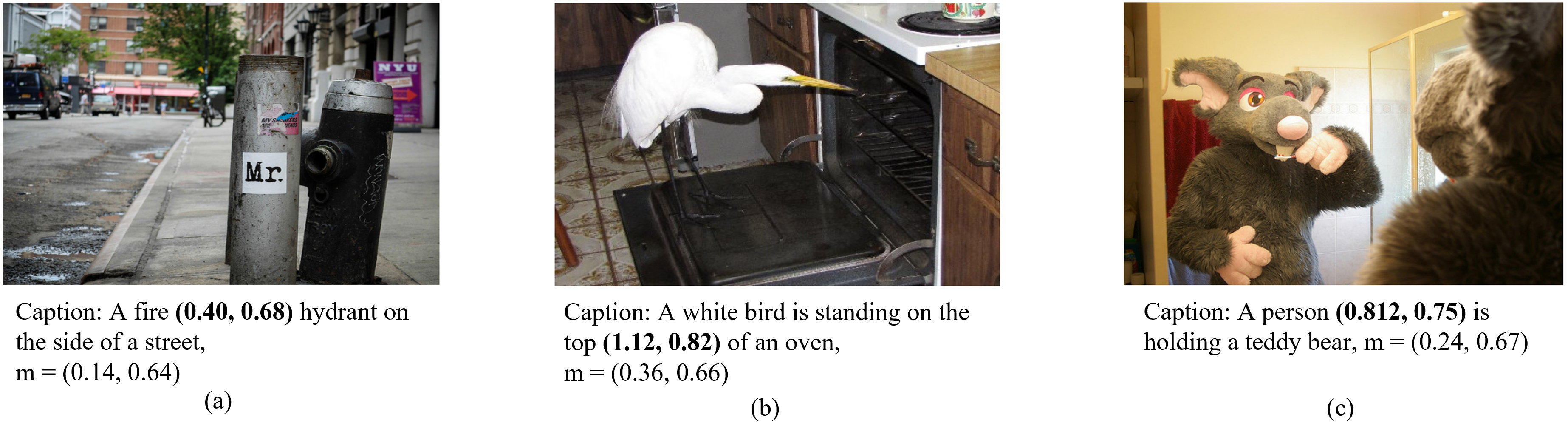}
    \caption{Image captioning results on OK-VQA dataset. Values in bracket are aleatoric uncertainty and epistemic uncertainty, respectively, and $m$ represents an average aleatoric uncertainty and an average epistemic uncertainty in a sentence, respectively.} \label{fig:uncertainty}
\end{figure*}

Table \ref{tab:2} shows the performances of the baseline model for caption generation with the OK-VQA dataset. When we compared the image caption performance of the Att2in, BuDn, and Transformer models with the OK-VQA dataset, overall, the Transformer model shows better performance than others, and our study uses the Transformer model for uncertainty modeling. Fig. \ref{fig:uncertainty} shows aleatoric uncertainty and epistemic uncertainty of the word in the generated caption, and the word for uncertain actions and unusual objects in the image shows higher uncertainty than the average uncertainty of the sentence.  

\begin{figure}
    \centering
    \includegraphics[width=8cm]{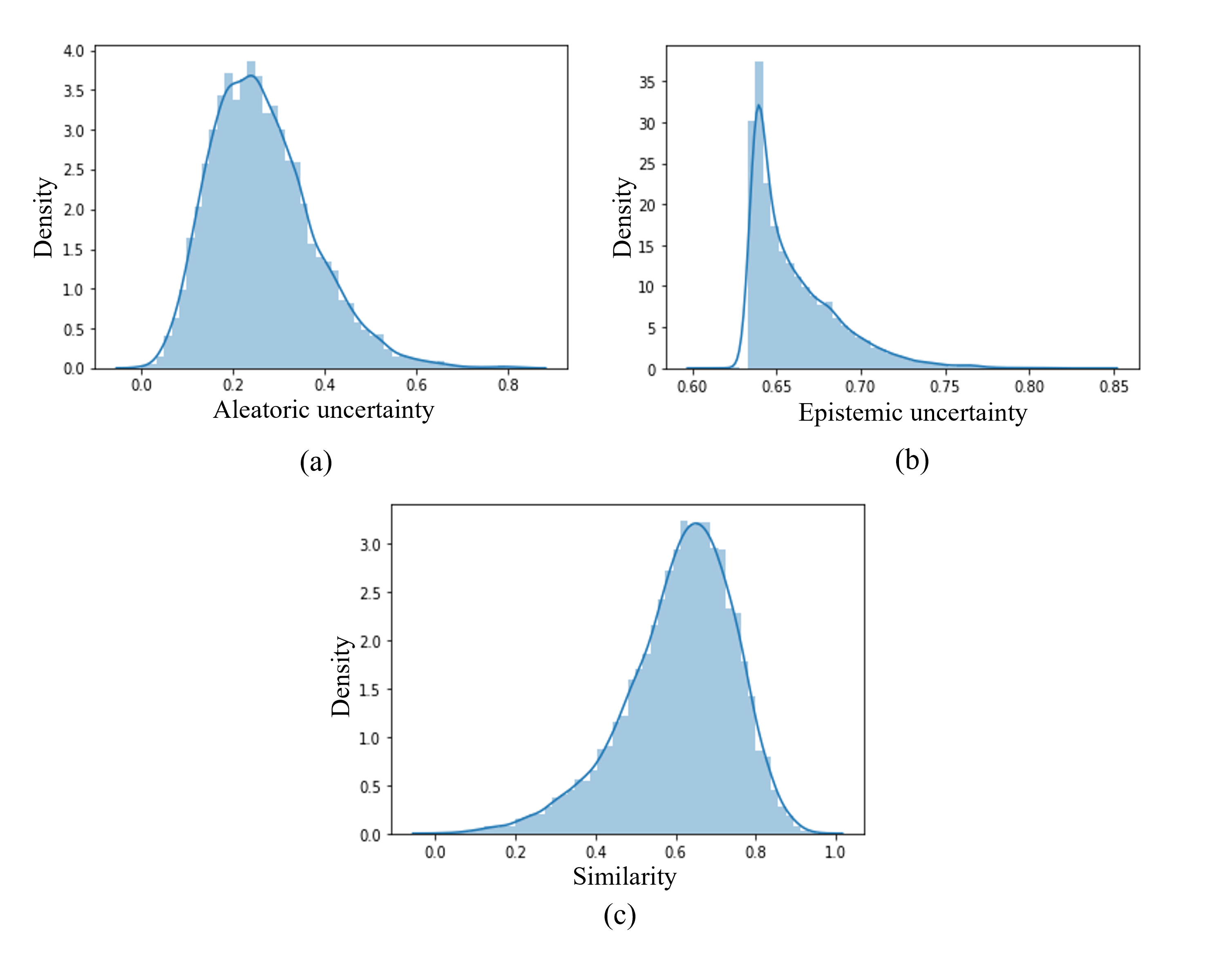}
    \caption{The distribution of the uncertainty of generated caption and the similarity between the caption and the ground-truth caption.}
    \label{fig:dist}
\end{figure}

\begin{figure}
    \centering
    \includegraphics[width=8.7cm]{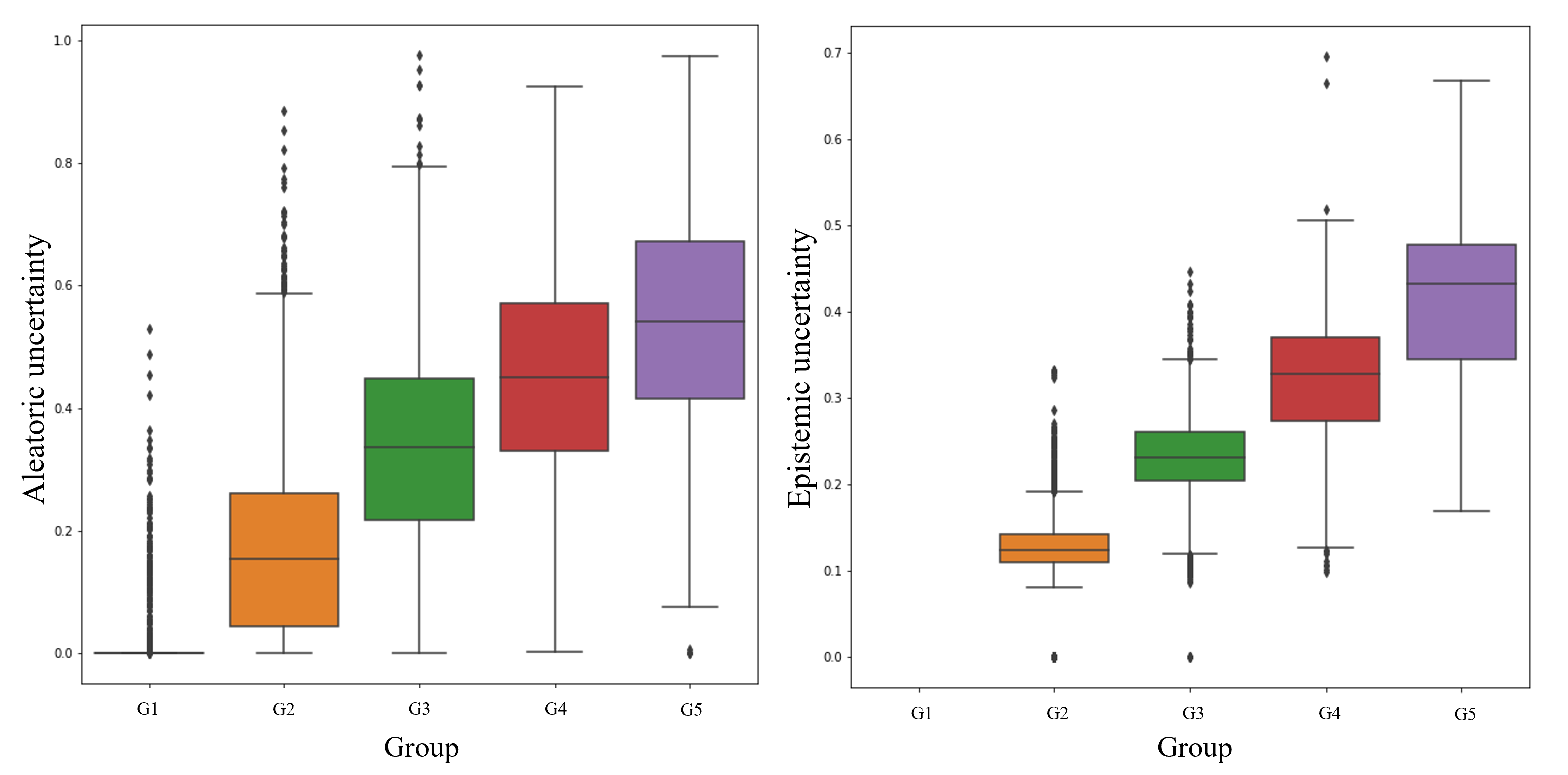}
    \caption{Boxplot of the uncertainty value according to the number of hallucinated objects.}
    \label{fig:boxplot}
\end{figure}

Table \ref{tab:3} illustrates the pearson correlation between uncertainty and caption similarity. The caption similarity and aleatoric uncertainty have a negative correlation of -0.1907, and the correlation between similarity and epistemic uncertainty is -0.1653. The correlation between aleatoric uncertainty and epistemic uncertainty shows a positive correlation, with a value of 0.4518. The correlation analysis indicates that there are relations between caption similarity and uncertainty, as we expected. In Fig. \ref{fig:dist}, the distributions of (a) aleatoric uncertainty and (b) epistemic uncertainty are right-skewed, while in (c) caption similarity distribution presents a left-skewed shape. Since there are extreme values in distributions, we believe that the semantic inconsistency can be identified with the uncertainties of the caption and the caption similarity. 
\begin{figure*}[ht]
    \centering
    \includegraphics[width=14cm]{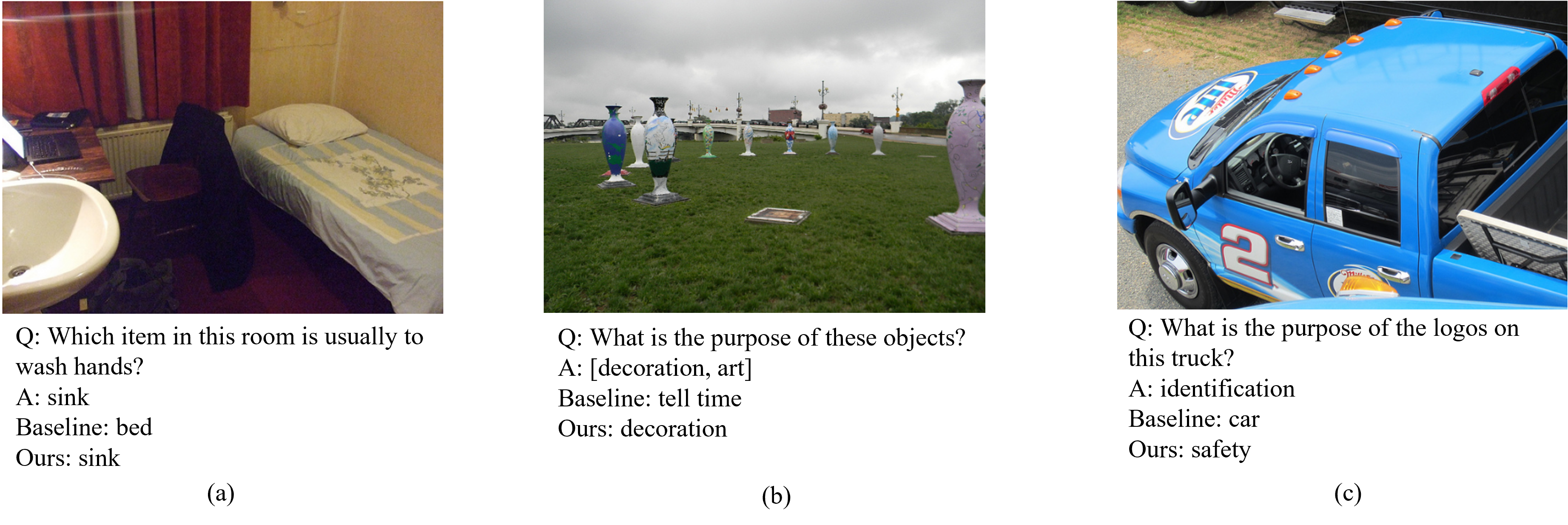}
    \caption{Comparison with the predicted answers of the proposed method and the baseline model on OK-VQA dataset.} \label{fig:vis}
\end{figure*}

We also analyzed the uncertainty relationship according to the number of hallucinated objects in the generated caption as shown in Fig. \ref{fig:boxplot}. In Fig. \ref{fig:boxplot}, the x-axis categories mean the following ranges, and the range means the ratio of the number of hallucinated words among the generated caption: 1) 0 $\leq$ G1 $<$ 0.2; 2)  0.2 $\leq$ G2 $<$ 0.4; 3)  0.4 $\leq$ G3 $<$ 0.6; 4)  0.6 $\leq$ G4 $<$ 0.8; 5)  0.8 $\leq$ G5 $\leq$ 1.0. The proportion of hallucinated objects of generated captions is calculated according to a synonym criteria of \cite{RN25}. After synonym filtering of the generated caption, the number of hallucinated objects in the generated caption is counted. We divide the ratio of the number of words of the hallucinated objects among the caption words into 5 groups. We calculate the average uncertainty of the caption over the average uncertainty of the hallucinated objects. As shown in Fig. \ref{fig:boxplot}, the more hallucinated objects in the caption, the higher aleatoric and epistemic uncertainty. We also performed a qualitative analysis, as shown in Fig. \ref{fig:uncertainty}. For the example shown in Fig. \ref{fig:uncertainty}, the generated caption contains uncertain words with higher aleatoric and epistemic uncertainty than $m$ the average aleatoric uncertainty and the average epistemic uncertainty in a sentence.

\subsection{KVQA with semantic inconsistency}
\subsubsection{Comparison with state-of-the-art approaches} We compare our proposed semantic inconsistency model with the following state-of-the-art approaches including those with pre-trained methods and a combination of graph-based and pre-trained methods: 1) BAN\cite{RN8}: Bilinear attention network which uses co-attention module with question features and image features from pre-trained models; 2) BAN+AN\cite{RN9}: The model incorporates the external knowledge into BAN by using ArticleNet; 3) BAN+KG-Aug\cite{RN10}: The model incorporates knowledge graph augmented model into BAN by using late augmentation scheme; 4) MUTAN\cite{RN12}: Multimodal tucker fusion network which focuses on image and textual features extracted from pre-trained models based on tucker decomposition; 5) MUTAN+AN\cite{RN9}: Similarly with BAN+AN, this method also incorporates the external knowledge into MUTAN by using ArticleNet; 6) KA\cite{RN14}: The model uses image features, question features, and concept graphs with the external knowledge; 7) KRISP\cite{RN2}: The model integrates image-text representation extracted from the BERT-based model and graph information based on external knowledge. In general, the methods using the knowledge information show better performance. As shown in Table \ref{tab:5}, the model with both explicit knowledge, implicit knowledge, and semantic inconsistency measure achieves the state-of-the-art performance. 
\subsubsection{Ablation study} An ablation study is performed with three values of caption similarity, aleatoric uncertainty, and epistemic uncertainty with the weights in Eq. (\ref{eqn:7}). In Table \ref{tab:4}, the baseline model that makes use of both explicit and implicit knowledge shows an accuracy of 31.15\%.  When caption similarity is added, the accuracy increases by 0.4\%. In addition, when aleatoric and epistemic uncertainty are added, respectively, it shows further improvement. Also, when the similarity and epistemic uncertainty are added, the accuracy increases by 0.49\%. The best performance of 32.45\% in accuracy is achieved when the caption similarity and the aleatoric uncertainty are concatenated. As shown in Table \ref{tab:3}, since the similarity and aleatoric uncertainty has a higher correlation than between the similarity and epistemic uncertainty, the concatenated model seems to provide the best performance. When the three values of caption similarity, aleatoric uncertainty, and epistemic uncertainty are used together, the accuracy is 31.19\%, which is only slightly better than the baseline. These results indicate that the caption similarity captures the semantic inconsistency relatively well and when the value which has a high correlation with the similarity is given to the model, it can predict correct answers better. 

\begin{table}[ht]
\caption{An ablation study of the external knowledge assimilation methods with the OK-VQA dataset.} \label{tab:4}
\begin{center}
\begin{adjustbox}{width=4.2cm,center}
\begin{tabular}{cc}
\toprule
\multicolumn{1}{c}{\bf Model}  &\multicolumn{1}{c}{\bf Accuracy}  \\ \toprule
Baseline     &  31.15   \\ \midrule
Baseline +\\ $sim^{cap}$     &  31.55    \\ \midrule
Baseline +\\ $uncertainty^{al}$      &  31.28    \\ \midrule
Baseline +\\ $uncertainty^{ep}$      &  31.93    \\ \midrule
Baseline +\\ $sim^{cap}$ +\\ $uncertainty^{ep}$     &  31.64    \\ \midrule
Baseline +\\ $sim^{cap}$ +\\ $uncertainty^{al}$      &  \bf 32.45    \\ \midrule
Baseline +\\ $sim^{cap}$ +\\ $uncertainty^{ep}$ +\\ $uncertainty^{al}$  &  31.19    \\\bottomrule
\end{tabular}
\end{adjustbox}
\end{center}
\end{table}

\begin{table}[ht]
\caption{Results with the OK-VQA dataset, comparing our work with the state-of-the-art approaches. * represents results from a re-implementation with the author's code and parameter setting using three experiments.} \label{tab:5}
\begin{center}
\begin{adjustbox}{width=4.9cm,center}
\begin{tabular}{>{\centering\arraybackslash}p{0.5\textwidth}>{\centering\arraybackslash}p{0.1\textwidth}}
\toprule
\multicolumn{1}{c}{\bf Model}  &\multicolumn{1}{c}{\bf Accuracy}  \\ \toprule
Q-Only    &  14.93   \\ \midrule
BAN \cite{RN8}     &  25.17    \\ \midrule
BAN + AN \cite{RN9}    &  25.61    \\ \midrule
MUTAN \cite{RN12}     &  26.41    \\ \midrule
BAN + KG-Aug \cite{RN10}    &  26.71    \\ \midrule
MUTAN + AN \cite{RN9}    &  27.84    \\ \midrule
KA \cite{RN14}   &  29.03    \\ \midrule
KRISP* \cite{RN2}     &  31.15    \\ \midrule
Ours  &  \bf 32.45    \\\bottomrule
\end{tabular}
\end{adjustbox}
\end{center}
\end{table}

\subsubsection{Qualitative results} We also present a qualitative analysis of the model in Fig. \ref{fig:vis}. We compare the prediction from our model with the baseline's. For (a) and (b), our model selects the correct answer. In addition, for (c) our model predicts an answer that is more similar to the correct answer than the baseline model. Also, in Fig. \ref{fig:vis}, the proposed method predicts correct answers even when the image shows a part of the sink (a), and with an unusual combination of objects and the background (b).

\section{Conclusion and future work} \label{con}
In this study, we propose a novel semantic inconsistency measure through uncertainty modeling and semantic similarity for KVQA that can make use of diverse KBs more effectively. As KBs are often incomplete or incompatible with the given problem, the use of knowledge should be moderated. With the proposed model, we achieve the state-of-the-art results on KVQA. As a future work, we plan to further explore diverse ways of using KBs based on the characteristics of the KB and the given problem. 

\section{Acknowledgements} 
This research was supported by the MSIT (Ministry of Science and ICT), Korea, under the ITRC (Information Technology Research Center) support program (IITP-2022-2020-0-01789) supervised by the IITP (Institute for Information \& Communications Technology Planning \& Evaluation).

\bibliographystyle{unsrt}
\bibliography{IJCNN} 

\end{document}